\documentclass[10pt,twocolumn,letterpaper]{article}

\usepackage[]{cvpr}      % To produce the REVIEW version
\usepackage{times}
\usepackage{epsfig}
\usepackage{graphicx}
\usepackage{amsmath}
\usepackage{amssymb}
\usepackage{booktabs}
\usepackage{bbding}
\usepackage{multirow}
\usepackage{makecell}
\usepackage{caption}
\usepackage{pifont}

\usepackage[pagebackref,breaklinks,colorlinks=true,citecolor=RoyalBlue,linkcolor=magenta]{hyperref}

\usepackage[dvipsnames]{xcolor}         % colors
\usepackage{colortbl}
\usepackage{cuted}

\definecolor{baselinecolor}{gray}{.9}
\newcommand{\baseline}[1]{\cellcolor{baselinecolor}{#1}}

\newcolumntype{x}[1]{>{\centering\arraybackslash}p{#1pt}}
\newcolumntype{y}[1]{>{\raggedright\arraybackslash}p{#1pt}}
\newcolumntype{z}[1]{>{\raggedleft\arraybackslash}p{#1pt}}
\newlength\savewidth

\usepackage[capitalize]{cleveref}
\crefname{section}{Sec.}{Secs.}
\Crefname{section}{Section}{Sections}
\Crefname{table}{Table}{Tables}
\crefname{table}{Tab.}{Tabs.}
\usepackage[dvipsnames]{xcolor}

\definecolor{cvprblue}{rgb}{0.21,0.49,0.74}
\title{Adaptive Fusion of Single-View and Multi-View Depth for Autonomous Driving}

\author{
JunDa Cheng$^{1,2}$\footnotemark[1],
~~Wei Yin$^{2}$\footnotemark[1],
~~Kaixuan Wang$^{2}$,
~~Xiaozhi Chen$^{2}$,
~~Shijie Wang$^{1}$,
~~Xin Yang$^{1}$\footnotemark[2]\\
[2mm]
$^1$~Huazhong University of Science and Technology \quad $^2$~DJI Technology \\
{\tt\small \{Junda Cheng, sjw, xinyang2014\}@hust.edu.cn, yvanwy@outlook.com}}

\begin{document}
\maketitle

\begin{strip}
  \centering
    \includegraphics[width=0.87\linewidth,page=1]{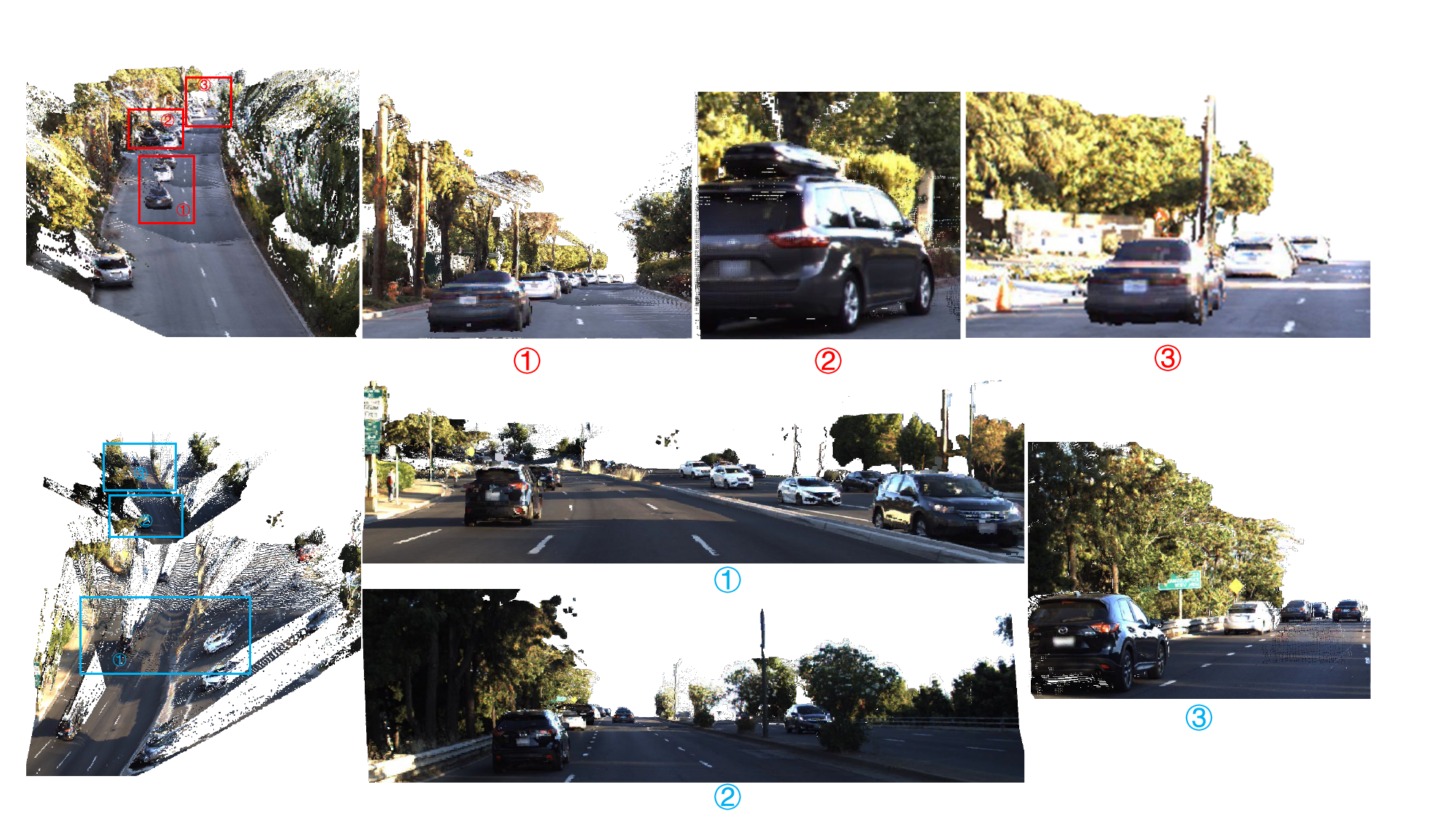}
    \vspace{-5pt}
    \captionof{figure}{%Point cloud 
    Visualization of reconstructed 3D point clouds of DDAD~\cite{godard2019digging} scenes. We fuse the results of 10 frames (including the dynamic object cars) and zoom in on some details for visualization. %It can be seen 
    It shows that %the results of our point cloud are 
    our method can achieve high-quality results on %robust in 
    both static and dynamic parts.%areas.
    }
    \label{fig:point}
    \vspace{-5pt}
\end{strip}

\renewcommand{\thefootnote}{\fnsymbol{footnote}}
\footnotetext[1]{Equal contribution.}
\footnotetext[2]{Corresponding author.}

\begin{abstract}
% \vspace{-10pt}
Multi-view depth estimation has achieved impressive performance over various benchmarks. However, almost all current multi-view systems rely on given ideal camera poses, which are unavailable in many real-world scenarios, such as autonomous driving. In this work, we propose a new robustness benchmark to evaluate the depth estimation system under various noisy pose settings. Surprisingly, we find current multi-view depth estimation methods or single-view and multi-view fusion methods will fail when given noisy pose settings. To address this challenge, we propose a single-view and multi-view fused depth estimation system, which adaptively integrates high-confident multi-view and single-view results for both robust and accurate depth estimations. The adaptive fusion module performs fusion by dynamically selecting high-confidence regions between two branches based on a wrapping confidence map. Thus, the system tends to choose the more reliable branch when facing textureless scenes, inaccurate calibration, dynamic objects, and other degradation or challenging conditions. Our method outperforms state-of-the-art multi-view and fusion methods under robustness testing. Furthermore, we achieve state-of-the-art performance on challenging benchmarks (KITTI and DDAD) when given accurate pose estimations.  Project website: \textcolor{magenta}{https://github.com/Junda24/AFNet/}.
\end{abstract}

% \vspace{-5pt}

% \input{sec/0_abstract}    
% \input{sec/1_intro}
% \input{sec/2_formatting}
% \input{sec/3_finalcopy}
\section{Introduction}
%%%%%%%%% BODY TEXT
% from kaixuan
Depth estimation from images is a long-standing problem in computer vision with wide applications. For vision-based autonomous driving systems, % applications. Perceiving depth in an autonomous driving scene is 
perceiving depth is an indispensable module %helpful 
for understanding the correlation of road objects and modeling 3D environment maps.
% environment layout.
Since deep 
neural networks are applied to solve various vision problems,  %has been the dominant solutions for depth estimation in recent years. 
CNN-based methods~\cite{yao2018mvsnet, chen2019point,gu2020cascade,yin2021virtual, bhat2021adabins, yuan2022new, Yin2019enforcing, cheng2024coatrsnet,xu2022attention,xu2023accurate,xu2023iterative,xu2023cgi, cheng2022region} have dominated various depth benchmarks. %There are two main streams for the depth perception: 

According to the input formats, they are mainly categorized into multi-view depth estimation~\cite{zbontar2015computing, yao2018mvsnet, chen2019point,gu2020cascade,long2021multi,nie2019multi,yao2019recurrent,yu2020fast} and single-view depth estimation~\cite{xu2017multi, hao2018detail, xu2018structured, hu2019revisiting, lee2019big, huynh2020guiding}. Multi-view methods estimate depth under the assumption that given correct depth, camera calibration, and camera poses, the pixels should be similar across views. They rely on epipolar geometry to triangulate high-quality depth. However, the accuracy 
and robustness 
of multi-view methods %will be affected by 
heavily rely on the geometric configuration of the camera and the %quality of the correspondences among views
correspondences matching among views. 
First, the camera is required to translate sufficiently for triangulation. %For an autonomous car, 
In autonomous driving scenarios, %the camera can be static or rotate without translation, such that triangulation is not possible. 
the car may stop at traffic lights or turn around without moving forward, which causes failure triangulation. 
Furthermore, the multi-view methods %can not work well on
suffer from dynamic objects and textureless regions, which are %common scenarios 
ubiquitous in autonomous driving scenarios. %Furthermore,
Another problem is SLAM pose optimization on moving vehicles. 
Noises are inevitable in existing SLAM methods, not to mention challenging and inevitable situations. For example, one robot or autonomous car can be deployed for years without re-calibration, causing noisy poses. 
%carefully calibrated intrinsic or accurate camera pose estimations are unavailable in many situations, which are vital for the accurate prediction of multi-view methods. One robot or autonomous car can be deployed for years without re-calibration and the noise is inevitable in existing SLAM methods to retrieve
%pose information when the vehicle is moving. %These limitations encourage the researcher to develop single-view approaches that require only one image and no calibration information.
In contrast, as single-view methods~\cite{xu2017multi, hao2018detail, xu2018structured, hu2019revisiting, lee2019big, huynh2020guiding} rely on the semantic understanding of the scene and the perspective projection cues, they are more robust to textureless regions, dynamic objects, and not rely on camera poses. However, its performance is still far from the multi-view methods because of scale ambiguity. Here, we tend to think about if both methods' benefits can be well combined for robust and accurate monocular video depth estimation in autonomous driving scenarios.

Although the fusion-based systems have been explored in previous work~\cite{bae2022multi,facil2017single}, they all assume %that the ideal camera poses can be obtained for the multi-view system
ideal camera poses. %Thus, 
The consequence is %the performance of the fusion system 
fusion system's performance is %will be 
even worse than single-view depth estimation %when 
given noise poses. %On the contrary, 
To solve this problem, %in this paper, 
we propose a novel adaptive fusion network to exploit the advantages of both the multi-view and single-view methods and mitigate their disadvantages, which maintain high precision and also improve the robustness of the system under noisy poses. 
%More 
Specifically, %our network has two branches, 
we propose a two-branch network, i.e. one targets monocular depth cues, while the other one leverages the multi-view geometry. Two branches both predict a depth map and a confidence map.
%one is the single-view branch, and the other is the multi-view branch. 
To supplement the semantic cues and edge details lost in the cost aggregation of the multi-view branch, we first fuse the monocular features in the decoder part. 
%Firstly, we carry out fusion at the feature level, i.e., conduct fusion of single-view feature and multi-view feature in the decoder part to supplement semantic and edge details lost in the cost aggregation of the multi-view branch. 
%
We further design an adaptive fusion (AF) module to achieve the final merged depth.  Apart from the predicted confidence, we design another wrapping confidence map by performing the multi-view texture consistency check with the predicted depth and provided camera poses. 
%warping confidence map obtained by the mutual projection of multi-view images based on the relative pose and depth of the multi-view branch.
%
%Then we fuse at the depth level, which uses the proposed adaptive fusion (AF) module to perform fusion through three confidence maps, one is the confidence of the single-view branch, one is the confidence of the multi-view branch, and the last one is the warping confidence map obtained by the mutual projection of multi-view images based on the relative pose and depth of the multi-view branch. 
We can easily notice the inconsistency projection 
%The projection relationship is not satisfied 
when the pose or the depth is inaccurate, %and when 
or dynamic %the 
objects appear. % is dynamic, it will probably be projected to visually dissimilar pixels in the other images. 
%By combining the warping confidence map with the predicted confidence maps,% of the two branches, the relatively accurate depth per pixel between predictions of the two branches can be fused as the final depth.
By using such proposed confidence maps to perform the pixel-wise fusion,  we can achieve a much more robust depth finally. Our contributions are summarized below.

% To further improve the accuracy of the network when the pose fails completely, we propose a pose correction module, which can adaptively replace the input pose with the predicted pose by Posenet~\cite{kendall2015posenet} for multi-view estimation.
\begin{itemize}
    \setlength{\parskip}{0pt}
    \setlength{\parsep}{0pt}
    \item We propose AFNet to adaptively fuse the single-view and multi-view depth for more robust and accurate depth estimation. %for multi-view depth estimation in the wild, 
    It %achieving 
    achieves the state-of-the-art performance on both KITTI~\cite{geiger2013vision} and DDAD~\cite{godard2019digging} datasets.
    % which are typical autonomous driving scenarios.
    \item We are the first to propose the multi-view and single-view depth fusion network for alleviating the defects of the existing multi-view methods which will fail under noisy pose. We propose a new robustness testing benchmark to explore the effectiveness of the multi-view methods given noisy poses to prove this. Our method outperforms all other classical multi-view methods under noisy poses.
    % \item We design the pose correction module to detect and replace noisy poses, which can raise the upper limit of network accuracy in extreme pose degradation.
    \item Our AF module can improve the performance of the dynamic object regions which cannot be well processed by the classical multi-view depth estimation methods.
\end{itemize}
%-------------------------------------------------------------------------
\section{Related work}
\label{sec:Related work}

%\textbf{Single-View depth.} 
\textbf{Monocular depth estimation.} %Single-view 
Monocular depth estimation is a classical problem in computer vision. Recent CNN-based methods~\cite{yin2021virtual, bhat2021adabins, yuan2022new, Yin2019enforcing, li2023, li2023learning} mainly formulate it as a per-pixel classification~\cite{bhat2021adabins, fu2018deep, Yin2019enforcing} or regression problem~\cite{lee2019big, hao2018detail, laina2016deeper, xu2018structured}. To boost the performance, some methods~\cite{yuan2022new, Qi_2018_CVPR} propose to aggregate stronger vision features, some methods propose various losses~\cite{bhat2021adabins, yin2021virtual, Yin2019enforcing}, and some also propose to leverage mix-data training~\cite{yin2020diversedepth, yin2022towards, eftekhar2021omnidata}. %Although they demonstrate good performance in specific environments, the application domain of these methods is limited by the training dataset. The scale ambiguity is another reason that limits the usability.
Although their performance on various benchmarks has been improved continuously, the state-of-the-art accuracy is still far from the multi-view geometry-based methods. In our work, we integrate the single-view depth estimation module in our system because of such methods' robustness to low-texture regions and dynamic objects. 

\textbf{Multi-View depth estimation.} A variety of works have been proposed to estimate the depth based on multi-view observation with known intrinsics and camera poses. \cite{zbontar2015computing} is the first one to bring the power of feature learning into multi-view stereo, but they process the matching costs with the traditional aggregation method. ~\cite{yao2018mvsnet} proposed to first construct a differentiable cost volume and then use 3D CNNs to regularize the cost volume obtaining the most advanced accuracy at that time. 
% Following that, many cost volume-based methods~\cite{chang2018pyramid,gu2020cascade,yao2019recurrent,cheng2024coatrsnet,xu2022attention,xu2023accurate,xu2023iterative,xu2023cgi, cheng2022region} have been proposed. 
Most recent state-of-the-art methods follow such a paradigm ~\cite{chen2019point,long2021multi,nie2019multi,yu2020fast}. But these methods have a strong dependency on high-parallax motion and heterogeneous-texture scenes for high accuracy, and can't handle dynamic objects, etc. In addition, these methods require accurate relative pose between multiple frames and have poor anti-interference ability to pose noise.

\begin{figure*}
\centering
\includegraphics[width=0.88\textwidth]{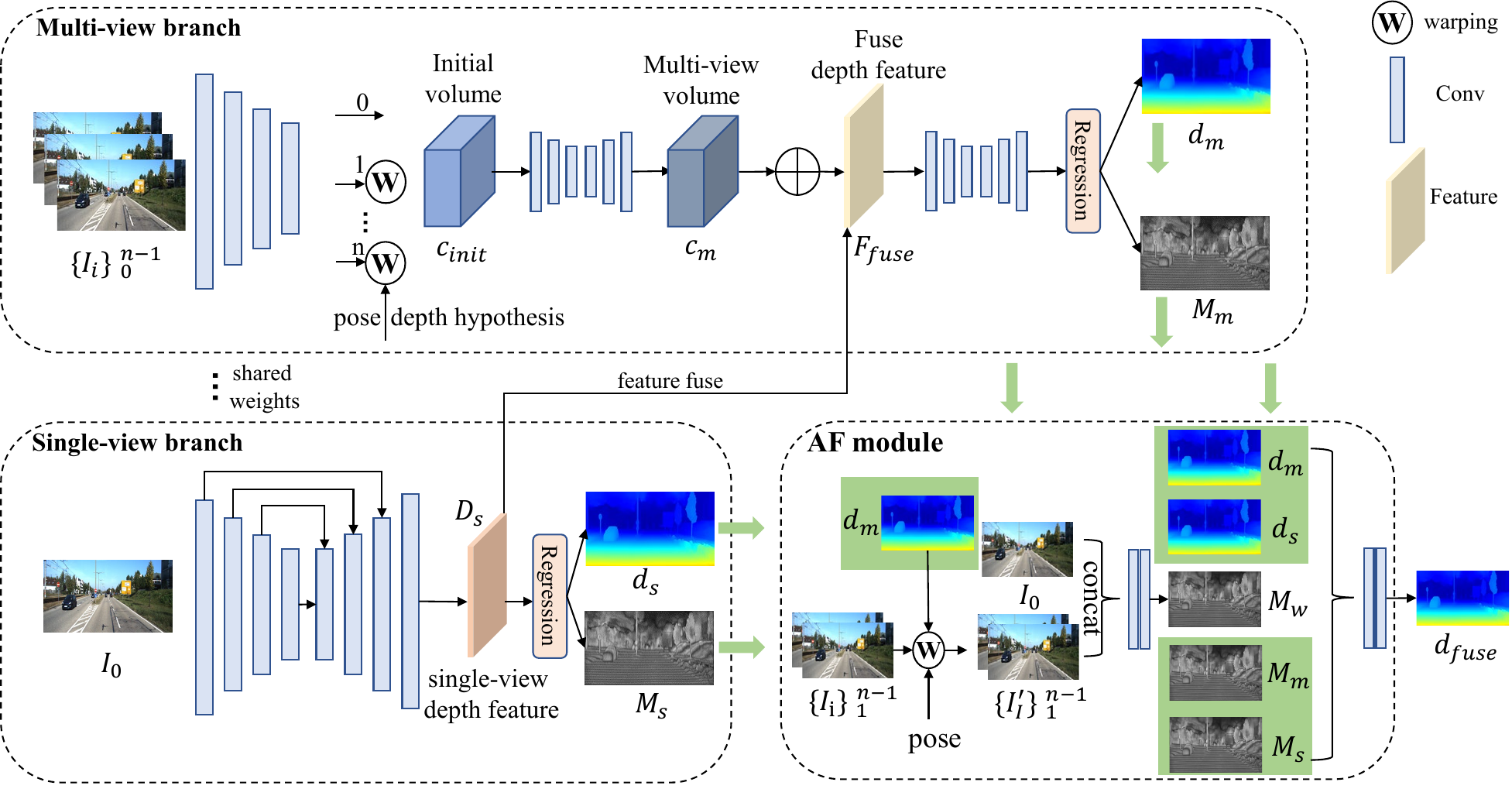} % Reduce the figure size so that it is slightly narrower than the column.
\vspace{-5pt}
\caption{Overview of the AFNet, which consists of three parts: single-view branch, multi-view branch, and the adaptive fusion (AF) module. Two branches share the feature extraction network and have their own prediction and confidence map, i.e. $\boldsymbol{d}_{s}$, $\boldsymbol{M}_{s}$, $\boldsymbol{d}_{m}$ and $\boldsymbol{M}_{m}$, and then fused by the AF module to obtain the final accurate and robust prediction $\boldsymbol{d}_{fuse}$. The green background in AF module represents the outputs of the single-view branch and multi-view branch.}
\label{fig:network1}
\vspace{-10pt}
\end{figure*}

% \section{Method}
% \label{sec:method}

% \begin{figure}
% \centering
% \includegraphics[width=1.0\linewidth]{./figure/network2.png} % Reduce the figure size so that it is slightly narrower than the column.
% \caption{The structure of our proposed adaptive fusion module.
% }
% \label{fig:network2}
% \vspace{-15pt}
% \end{figure}

\textbf{Single and multi-View depth fusion.}
There are also some methods~\cite{facil2017single,yang2022mvs2d,bae2022multi} for integrating single-view and multi-view prediction to exploit the properties of these two methods. But they basically focus on improving accuracy and efficiency. \cite{facil2017single} improves the accuracy by combining the local structure of single-view and the reliable prediction of multi-view in high-parallax and high-gradient regions. ~\cite{bae2022multi} first predicts a depth through a single-view network, then builds a thin cost volume based on this initial depth to reduce the high computation overhead and achieve a certain accuracy improvement. However, the above two methods all ignore a problem: multi-view branches are highly dependent on the accuracy of the pose. With noisy poses, the accuracy decreases seriously, even worse than with single-view methods.
% Once the pose noise is too large, the accuracy of multi-view branches will be greatly reduced, even lower than that of single-view branches, resulting in worse results after fusion. 
~\cite{yang2022mvs2d} proposed an Epipolar Attention Module to fuse single-view and multi-view geometric information, and installing attention modules at different resolutions can alleviate the problem of pose inexact to a certain extent. However, this approach can only slightly alleviate the pose noise problem, the wrong matching information extracted through the wrong poses is still integrated into the single-view branch, leading to the decay of accuracy.

% To sum up, we propose AFNet, which not only improves the accuracy, but also adaptively determines whether the current prediction of multi-view branch is reliable through an adaptive fusion module, only the reliable multi-view prediction and single-view will be integrated. When the pose is accurate, our accuracy can exceed the current multi-view sota methods, and when the pose fails, our method can at least obtain single-view accuracy.
Thus, we propose AFNet which adaptively combines the reliable results of the two branches through the AF module to enhance the robustness of the system under noisy poses.

\vspace{-5pt}
\section{Method}
\label{sec:method}

\subsection{Method Overview} \label{sec:method_overview}

We aim to estimate the depth $\boldsymbol{d} \in \mathcal{R}^{H \times W}$ for a reference image $\boldsymbol{I}_{0} \in \mathcal{R}^{H \times W \times 3}$, given $n$-1 source images $ \left\{\boldsymbol{I}_{i}\right\}_{i=1}^{n-1}$,
% captured at nearby views 
%with known intrinsic 
camera 
intrinsics and camera poses. Figure~\ref{fig:network1} overviews our AFNet, which %mainly 
consists of three parts, i.e.  %single-view branch, multi-view branch (Section~\ref{sec:single and multi}) 
single-view depth module, multi-view depth module, and adaptive fusion module. 
% We first extract the features of the reference image and the source image through a feature extraction network with shared parameters. The single-view branch directly inputs the extracted features of the reference image into a lightweight decoder network to obtain the single-view depth. The multi-view branch first uses the features of the reference image and source image to construct the cost volume, and then sends it to decoder for aggregation. In addition, the depth feature of single-view is integrated into the decoder part to obtain the depth of the multi-view branch. 
% Both of two branches have their own depth prediction, and then fused by the adaptive fusion module to obtain the final accurate and robust prediction. 
Furthermore, 
% when the provided poses include much noise, 
we propose a pose correction module to ensure robustness under large noise poses, the details can be found in the supplementary materials. 
%We also propose a pose correction module to further improve the accuracy when the pose is severely interfered (Section~\ref{sec:posecorrect}).
% We also propose a pose correction module which can independently identify when the input pose noise is too large, and then replace the input noisy pose with the pose predicted (Section~\ref{sec:posecorrect}).

\subsection{Single-view and Multi-view Depth Module} \label{sec:single and multi}
%\textbf{Feature extractor.} 
% Given $n$ input images $\left\{I_{i}\right\}_{i=0}^{n-1}$ of size $W \times H$.
% , we use $I_{0}$ and $\left\{I_{i}\right\}_{i=1}^{n-1}$ to denote reference and source images respectively.
In our system, we use %a lightweight convolutional encoder of 
ConvNeXt-T~\cite{liu2022convnet} as the backbone to extract 4 scales features $\boldsymbol{{F}}_{i,l}$ ($l$ = 1, 2, 3, 4), %. We attain features at 4 scale levels. denote the feature of the $i$-th image at level $l$ by ${F}_{i,l}$.
where $i$ is the index of the image and $l$ is the scale. The extracted 4-scale feature dimensions are $C = 96, 192, 384, 768$ respectively.

%The features ${F}_{i,l}$ are stored at $1 / 2^{l+1}$ resolution and possess $C = 96, 192, 384, 768$ channels at level $l = 1, 2, 3, 4$ respectively.

\textbf{Single-view branch.} Following ~\cite{yang2022mvs2d}, we construct a multi-scale decoder to merge backbone features %After the multi-scale features ${F}_{0,l}$ of the reference image %is 
%being extracted, % through the feature extraction network, 
%we used the decoder structure of ~\cite{yang2022mvs2d} 
% denoted as D-Net for decoding
% . High resolution features are combined with the upsampled low level features to get single-view features $F_{s} \in \mathcal{R}^{H/4 \times W/4 \times 256}$ including multi-scale information. Then, the $F_{s}$ is sent to a lightweight U-NET network 
and obtain the depth feature $\boldsymbol{D}_{s} \in \mathcal{R}^{H/4 \times W/4 \times 257}$. By applying the softmax along the channel dimension for the first 256 channels of the $\boldsymbol{D}_{s}$, we get the depth probability volume $\boldsymbol{P}_{s} \in \mathcal{R}^{H/4 \times W/4 \times 256}$. The last channel of the feature serves as the confidence map $\boldsymbol{M}_{s} \in \mathcal{R}^{H/4 \times W/4}$ for the single-view depth. Finally, the single-view depth is computed by the soft-weighted sum. It is as follows.
\vspace{-5pt}
\begin{equation}
\boldsymbol{d}_{s}=\sum\limits_{d \in \mathbb{B} }d\cdot{p_{d}},
\label{equ:soft-argmin}
\end{equation}
% \vspace{-5pt}
where $\mathbb{B}$ denotes the bins uniformly sampled in the log space from $d_{min}$ to $d_{max}$, which represents the depth search range, $p_d$ denotes the corresponding probability in $\boldsymbol{P}_{s}$.

\textbf{Multi-view branch.} 
% After obtained the multi-scale feature 
%We use the same backbone as encoder and 
The multi-view branch shares the backbone with the single-view branch to extract features $\boldsymbol{{F}}_{i,l}$ for the reference and source images. 
% We adopted three 2d deconvolutions to gradually deconvolve the low-resolution features ${F}_{i,l}$ ($l$ = 2, 3, 4) to the quarter resolution and combine the initial quarter features ${F}_{i,1}$ and reduce the number of channels to reduce the computation, so that the combined features ${F}_{mvs} \in \mathcal{R}^{H/4 \times W/4 \times 32}$  can contain multi-scale information, 
We %adopted 
adopt deconvolutions to deconvolve the low-resolution features to the quarter resolution and %combined 
combine them with the initial quarter features $\boldsymbol{{F}}_{i,1}$, 
% and reduce channel number, 
which is used to construct the cost volume. %We follow Mvsnet~\cite{yao2018mvsnet}, 
By warping the source features into hypothesis planes of the reference camera follow~\cite{yao2018mvsnet}, the feature volumes are formed. 
% The homography transformation formula is as follows:
% \begin{equation}
% H_{i}(d)=K_{i} \cdot R_{i}\cdot (I - \frac{(t_{1}-t_{i})\cdot{n_{1}}^{T}}{d}) \cdot {R_{1}}^{T} \cdot {K_{1}}^{-1}
% \label{eq_mvsmapping}
% \end{equation}
% where $H_{i}(d)$ refers to the homography between the feature maps of the $i^{th}$ view and the reference feature maps at depth $d$.
% $K_{i},R_{i}, t_{i}$ refers to the camera intrinsics, rotations and translations of the $i^{th}$ view respectively, and $n_{1}$ denotes the principle axis of the reference camera. 
For robust matching information without much computation, we retain the channel dimension of the feature and construct 4D cost volume, and then reduce the channel number to 1 through two 3D convolution layers. The sampling method of the depth hypothesis is consistent with the single-view branch, but the sampling number is only 128, i.e. the initial cost volume $\boldsymbol{C}_{init} \in \mathcal{R}^{H/4 \times W/4 \times 128}$. 
% Then, in order to obtain multi-scale cost volume information, we use four 2D convolution layers with stride of 1,2,4,8 to obtain $C_{m, l} \in \mathcal{R}^{H/2^{l+1} \times W/2^{l+1} \times C_{l}}$ ($C_{l}$ = 96, 192, 384, 768 when $l$ switch 1 to 4) with the same size as ${F}_{i,l}$. Then we adopt the same D-Net as single-view branch for regularizing the multi-scale cost volume to 
Then we use a stacked 2D hourglass network for regularizing to obtain the final multi-view cost volume $\boldsymbol{C}_{m} \in \mathcal{R}^{H/4 \times W/4 \times 256}$. To supplement the rich semantic information of the single-view feature and the details lost due to cost regularization, we use a residual structure to combine the single-view depth feature $\boldsymbol{D}_{s}$ and cost volume to obtain the fuse depth feature $\boldsymbol{F}_{fuse}$ as follows:
\begin{equation}
\boldsymbol{{F}}_{fuse}=\text{Conv}\left\{\text{Concat}\left\{\boldsymbol{C}_{m},\boldsymbol{D}_{s}\right\}\right\} + \boldsymbol{C}_{m}.
\end{equation}
After using a 2D hourglass network for aggregating the $\boldsymbol{F}_{fuse}$, the subsequent operation is exactly the same as the single-view branch. 
% $D_{m} \in \mathcal{R}^{H/4 \times W/4 \times 257}$ is obtained through the same U-Net structure, and then 
% $P_{m} \in \mathcal{R}^{H/4 \times W/4 \times 256}$ and 
Confidence map of multi-view branch $\boldsymbol{M}_{m} \in \mathcal{R}^{H/4 \times W/4}$ and the final depth prediction $\boldsymbol{d}_{m}$ are obtained in the same regression way.

\subsection{Adaptive Fusion Module}
\label{sec:Adaptive}
% To obtain the final accurate and robust prediction, we designed adaptive fusion module to deal with degradation such as pose fail. It is well known that when the input pose is accurate, the accuracy of most areas of multi-view is higher than that of single-view. Therefore, we hope that the adaptive fusion module can independently judge whether the input pose is accurate. In addition, only the accurate region of multi-view prediction is kept, and the inaccurate region is filled by the prediction of single-view branch. Since the principle of monocular depth estimation does not involve pose and whether it is a dynamic object, the monocular depth estimation ensures the lower limit of our network prediction regardless of degradation conditions, and when the pose is accurate, we can reach the upper limit of SOTA multi-view methods. 
To obtain the final accurate and robust prediction, we design the AF module to adaptively select the most accurate depth between the two branches as the final output as shown in Figure~\ref{fig:network1}. We conduct fusion through three confidence maps, two of which are the confidence maps $\boldsymbol{M}_{s}$ and $\boldsymbol{M}_{m}$ generated by the two branches respectively (Section~\ref{sec:single and multi}), and the most critical one is the confidence map $\boldsymbol{M}_{w}$  generated by forward warping to judge whether the prediction of the muti-view branch is reliable. We use the camera pose and the multi-view depth $\boldsymbol{d}_{m}$ as the input to warping the source images to the reference camera space to obtain $\left\{\boldsymbol{I}_{i}^{'}\right\}_{i=1}^{n-1}$, and concat with $\boldsymbol{I}_{0}$. No matter the multi-view depth is inaccurate, the pose is noisy, or in the dynamic object area, the warping source images will be dissimilar with the corresponding pixels in the reference image. So we can build a warping confidence map $\boldsymbol{M}_{w} \in \mathcal{R}^{H/4 \times W/4}$ in this way as follows:
\begin{equation}
\boldsymbol{M}_{w}=\text{Conv}\left\{\text{Concat}\left\{\boldsymbol{I}_{0},\boldsymbol{I}_{1}^{'},\boldsymbol{I}_{2}^{'} \cdots  \boldsymbol{I}_{n-1}^{'}\right\}\right\}.
\end{equation}
% Then we have generated confidence maps $M_{s}$ and $M_{m}$ in single-view branch and multi-view branch respectively, which can also reflect the reliability of respective predictions.
$\boldsymbol{M}_{s}$ and $\boldsymbol{M}_{m}$ in single-view branch and multi-view branch reflect the overall matching ambiguity, while $\boldsymbol{M}_{w}$ reflects the subpixel accuracy. Thus, we take these three confidence maps as guidance for the fusion of single-view depth $\boldsymbol{d}_{s}$ and multi-view depth $\boldsymbol{d}_{m}$, and obtain the final fusion depth $\boldsymbol{d}_{fuse} \in \mathcal{R}^{H/4 \times W/4}$ by two 2D convolution layers.

\subsection{Loss Function}
The loss function during AFNet training is mainly composed of two parts, i.e. depth loss and confidence loss. Depth loss uses a simple L1 loss:
\begin{equation}
L_{d}=\left\|\boldsymbol{{d}}_{s}-\boldsymbol{{d}}_{gt}\right\|_{1} + \left\|\boldsymbol{{d}}_{m}-\boldsymbol{{d}}_{gt}\right\|_{1} + \left\|\boldsymbol{{d}}_{fuse}-\boldsymbol{{d}}_{gt}\right\|_{1}.
\end{equation}

For confidence loss, to prevent outliers from interfering with training, we first calculate the valid mask as follows:
\begin{equation}
\begin{split}
\boldsymbol{\Omega}_{s} =| \boldsymbol{d}_{s} - \boldsymbol{d}_{gt}| < \boldsymbol{d}_{gt}. \\
\boldsymbol{\Omega}_{m} = |\boldsymbol{d}_{m} - \boldsymbol{d}_{gt}| < \boldsymbol{d}_{gt}.
\end{split}
\end{equation}

The final confidence loss is calculated as follows:
\begin{equation}
\begin{split}
L_{c} = \frac{1}{N_{s}} \sum\limits_{p \in \boldsymbol{\Omega}_{s}} |\boldsymbol{M}_{s}(p) - (1 - |\boldsymbol{d}_{s}(p) - \boldsymbol{d}_{gt}(p)| / \boldsymbol{d}_{gt}(p)) | \\
% \boldsymbol{L_{c}} = M_{s}[m_{s}] - \left\{1 - (\boldsymbol{|}d_{s}[m_{s}] - d_{gt}[m_{s}]\boldsymbol{|})/d_{gt}[m_{s}]\right\} \\
+ \frac{1}{N_{m}} \sum\limits_{q \in \boldsymbol{\Omega}_{m}} | \boldsymbol{M}_{m}(q) - (1 - \boldsymbol{|} \boldsymbol{d}_{m}(q) - \boldsymbol{d}_{gt}(q) \boldsymbol{|}  / \boldsymbol{d}_{gt}(q)) |
% +  M_{m}[m_{m}] - \left\{1 - (\boldsymbol{|}d_{m}[m_{m}] - d_{gt}[m_{m}]\boldsymbol{|})/d_{gt}[m_{m}]\right\}
\end{split}
\end{equation}
$N_{s}$ and $N_{m}$ respectively represent the total number of valid points in $\boldsymbol{\Omega}_{s}$ and $\boldsymbol{\Omega}_{m}$. The total loss is the sum of the above two losses $L_{d}$ and $L_{c}$.

% When training the pose correction module, a pose constraint will be added to the output of the Posenet~\cite{kendall2015posenet}. First, 
% we convert the Ground Truth pose to Euler angles $\boldsymbol{r}_{i,gt}$ and translation matrixes $\boldsymbol{t}_{i,gt}$, and then calculate L1 loss with the prediction of $\boldsymbol{r}_{i,pred}$ and $\boldsymbol{t}_{i,pred}$:
% \begin{equation}
% \begin{split}
% L_{p} = \alpha * \sum_{i=1}^{i=n-1} \left\|\boldsymbol{r}_{i,pred} - \boldsymbol{r}_{i,gt}\right\|_{1} \\
% + \beta * \sum_{i=1}^{i=n-1} \left\|\boldsymbol{t}_{i,pred} - \boldsymbol{t}_{i,gt} \right\|_{1}.
% \end{split}
% \end{equation}

% $\alpha$ and $\beta$ are the respective weight coefficients.

% and then the translation matrix together constitutes the transformation matrix $T_{i,pred}$, i=1, 2 ,\cdots, n-1. 

%------------------------------------------------------------------------
\section{Experiment} \label{sec:experiment}
\renewcommand{\ttdefault}{ptm}
\newcommand{\tablestyle}[2]{\ttfamily\setlength{\tabcolsep}{#1}\renewcommand{\arraystretch}{#2}\centering\small}
% In this section, we experiment on two benchmarks, i.e., Dense Depth for Automated Driving (DDAD)~\cite{godard2019digging} and KITTI~\cite{geiger2013vision} to evaluate the superiority of our method.

%conducted ablation studies to prove the effectiveness of the components of our network. We also evaluate the proposed models on the Dense Depth for Automated Driving (DDAD) and KITTI benchmark to demonstrate the superiority of our method. Finally, we test the impact of pose containing noise on the depth accuracy on various classical multi-view methods and our AFNet, which proves the robustness of our method.
\begin{figure*}
\centering
\includegraphics[width=0.95\textwidth]{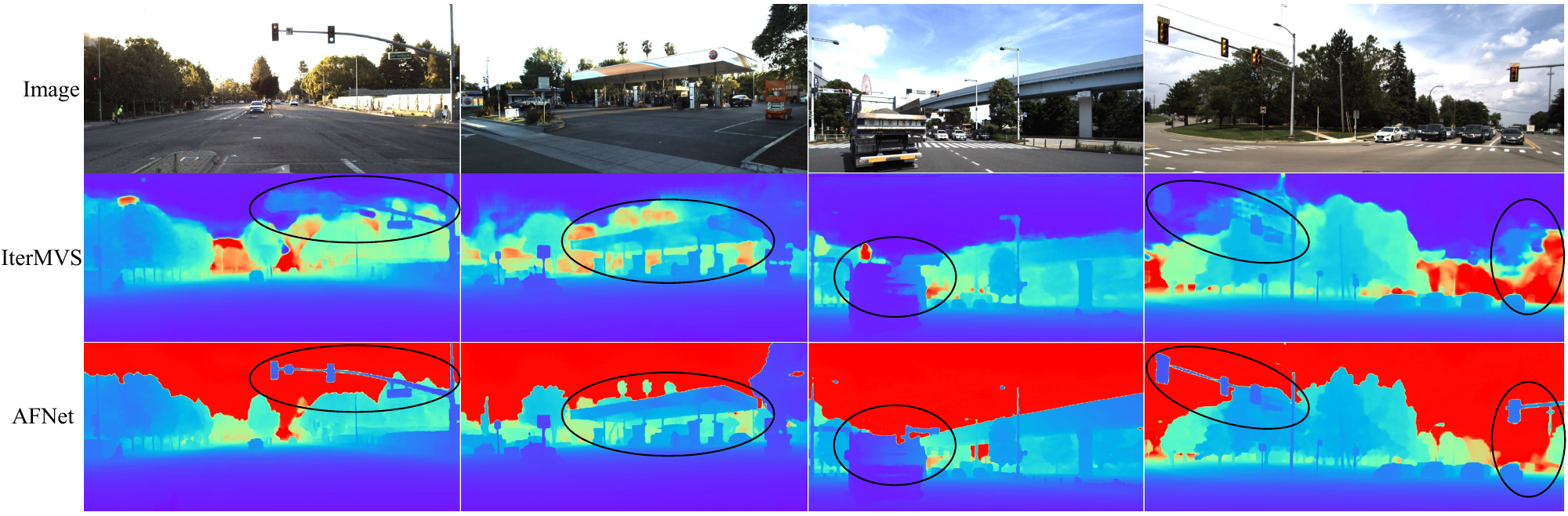} % Reduce the figure size so that it is slightly narrower than the column.
\vspace{-8pt}
\caption{Qualitative results on DDAD~\cite{godard2019digging} test set. Black ellipses highlight obvious improvements achieved by our method.
% , such as traffic lights, cars and gas stations.
}
\label{fig:ddadopt}
\vspace{-10pt}
\end{figure*}

\subsection{Datasets and Evaluation Metrics} \label{sec:Datasets}
\textbf{DDAD} (Dense Depth for Automated Drivin)~\cite{godard2019digging} is a new autonomous driving benchmark for dense depth estimation in challenging and diverse urban conditions. It is captured with 6 synchronized cameras and contains accurate ground-truth depth (across a full 360-degree field of view) generated from high-density LiDARs. It has 12650 training samples and 3950 validation samples in a single camera view, of which the resolution is 1936$\times$1216. The whole data from 6 cameras are used in training and testing.

\textbf{KITTI}~\cite{geiger2013vision} is a dataset that provides stereo images and corresponding 3D laser scans of outdoor scenes captured on a moving vehicle. The %RGB images have a 
resolution %of 
is around 1241$\times$376. We train and test our
method on KITTI
%, both using the Eigen split [12] and the official split. 
Eigen split~\cite{eigen2014depth}. 
%For all evaluations, depth accuracy is measured using the metrics defined in Table \ref{tab:metric}.
Metrics defined in Table~\ref{tab:metric} are used for evaluation. 

% \begin{small}
% \begin{table}[h] \footnotesize
% \centering

% \begin{tabular}{cc}
% \toprule
% AbsRel & $\frac{1}{N}\sum_{i}\frac{|d_i - d_i^{*}|}{d_i^{*}}$ \\
% SqRel & $\frac{1}{N}\sum_{i}\frac{(d_i - d_i^{*})^2}{ d_i^{*}}$ \\
% RMSE & $\sqrt{\frac{1}{N}\sum_{i}(d_i - d_i^{*})^2}$ \\
% \bottomrule
% \end{tabular}
% % \vspace{-5pt}
% \caption{Quantitative metrics for depth estimation. $d_i$ is the predicted depth; $d_i^*$ is the ground truth depth; $N$ corresponds to all pixels with the ground-truth label.}
% \label{tab:metric}
% % \vspace{-0.1in}
% \vspace{-15pt}
% \end{table}
\vspace{-5pt}
\begin{table}[h] \footnotesize
\centering

\begin{tabular}{c|c|c}
\toprule
AbsRel & SqRel & RMSE \\
% \hline
$\frac{1}{N}\sum_{i}\frac{(d_i - d_i^{*})^2}{ d_i^{*}}$ & $\frac{1}{N}\sum_{i}\frac{(d_i - d_i^{*})^2}{ d_i^{*}}$ & $\sqrt{\frac{1}{N}\sum_{i}(d_i - d_i^{*})^2}$ \\
\bottomrule
\end{tabular}
\vspace{-5pt}
\caption{Quantitative metrics for depth estimation. $d_i$ is the predicted depth; $d_i^*$ is the ground truth depth; $N$ corresponds to all pixels with the ground-truth label.}
\label{tab:metric}
% \vspace{-0.1in}
\vspace{-15pt}
\end{table}
% \end{small}

% \begin{table}[h]
% \centering
% \resizebox{\linewidth}{!}{%
% \begin{tabular}{cc|cc}
% \toprule
% AbsRel & $\frac{1}{N}\sum_{i}\frac{|d_i - d_i^{*}|}{d_i^{*}}$ & RMSE & $\sqrt{\frac{1}{N}\sum_{i}(d_i - d_i^{*})^2}$ \\
% SqRel & $\frac{1}{N}\sum_{i}\frac{(d_i - d_i^{*})^2}{ d_i^{*}}$ & AbsDiff & $\sqrt{\frac{1}{N}\sum_{i}|d_i - d_i^{*}|}$\\
% \bottomrule
% \end{tabular}
% }
% \caption{Quantitative metrics for depth estimation. $d_i$ is the predicted depth; $d_i^*$ is the ground truth depth; $N$ corresponds to all pixels with the ground-truth label.}
% \label{tab:metric}
% \vspace{-0.1in}
% \end{table}

% \begin{table}[h]
% \centering

% \begin{tabular}{cc}
% \toprule
% AbsRel & $\frac{1}{N}\sum_{i}\frac{|d_i - d_i^{*}|}{d_i^{*}}$\\ SqRel & $\frac{1}{N}\sum_{i}\frac{(d_i - d_i^{*})^2}{ d_i^{*}}$\\
% RMSE & $\sqrt{\frac{1}{N}\sum_{i}(d_i - d_i^{*})^2}$ \\
% \bottomrule
% \end{tabular}
% \caption{Quantitative metrics for depth estimation. $d_i$ is the predicted depth; $d_i^*$ is the ground truth depth; $N$ corresponds to all pixels with the ground-truth label.}
% \label{table:metric}
% \vspace{-0.1in}
% \end{table}

\subsection{Implementation Details} \label{sec:Implementation details}
We implement our methods with PyTorch~\cite{paszke2019pytorch} and perform experiments using NVIDIA RTX 3090 GPUs. We use AdamW optimizer~\cite{loshchilov2017decoupled} and schedule the learning rate using %1cycle 
one-cycle policy~\cite{smith2018super} with $lr_{max}$ = 1.0$\times$ $10^{-4}$. We trained 30 epochs on DDAD~\cite{godard2019digging} and 40 epochs on KITTI~\cite{eigen2014depth}. %For KITTI and DDAD, we use three images for multi-view, 
During training, our system consumes consecutive 3 frames as input, i.e, $n$=3. %For training the pose net of AFNet-pose, 
% The hyper-parameters in our pose net are $\left\{\alpha, \beta\right\} = \left\{300.0, 10.0\right\}$. 
For the details of multi-view setup, the depth hypothesis number of planes is 128, the weight of the images from different views is the same.

\begin{table*}[t]
  \centering
%   \tablestyle{10pt}{1.05}
  \begin{tabular}{c|c|ccc|ccc}
  \toprule
    \multirow{2}{*}{Type} & \multirow{2}{*}{Model} & \multicolumn{3}{c|}{DDAD~\cite{godard2019digging}}  & \multicolumn{3}{c}{KITTI~\cite{geiger2013vision}} \\ 
    \cmidrule{3-5} \cmidrule{6-8}
    & & AbsRel$\downarrow$  & SqRel$\downarrow$  & RMSE$\downarrow$  & AbsRel$\downarrow$  & SqRel$\downarrow$  & RMSE$\downarrow$ \\
    \hline
    \multirow{5}{*}{\makecell[c]{Single \\ View}} & Monodepth2~\cite{godard2019digging} & $0.194^*$ & $3.52^*$  & $13.32^*$  &  0.106 & 0.806 & 4.630 \\
    & FeatDepth~\cite{shu2020feature} & $0.189^*$  & $3.21^*$  & $12.45^*$  & 0.099 & 0.697 & 4.427 \\
    & DORN~\cite{fu2018deep}  & -  & -  & -  & 0.088 & 0.806 & 3.138 \\
    & BTS~\cite{lee2019big} & $0.169^*$  & $2.81^*$  & $11.85^*$  & 0.059 &  0.245 & 2.756 \\ 
    & AdaBins~\cite{bhat2021adabins} & $0.164^*$  & $2.66^*$  & $11.08^*$  & 0.058 & 0.190 & 2.360\\ 
    & Metric3D~\cite{yin2023metric3d} & $0.183^*$  & $2.92^*$  & $12.15^*$  & 0.053 & 0.174 & 2.243\\ 
    \hline
    \multirow{9}{*}{\makecell[c]{Multi \\ View}} & PMNet~\cite{wang2021patchmatchnet}  & $0.141^*$  & $2.23^*$  & $10.56^*$  & - & - & - \\ 
    & Deepv2d~\cite{teed2018deepv2d} & - & - & - & 0.091 & 0.582 & 3.644 \\
    & CasMVS~\cite{gu2020cascade} & $0.129^*$  & $2.01^*$  & $9.87^*$  & $0.066^*$  & $0.228^*$  & $2.567^*$  \\
    % & CascadeMVS~\cite{gu2020cascade} & 0.118 & 1.94 & 9.53 & 0.062 & 0.194 & 2.367\\
    % & Mvsnet~\cite{yao2018mvsnet}  & 0.135 & 2.18 & 10.72 & - & - & - & -\\
    % & SC-GAN~\cite{wu2019spatial} & - & - & - & 0.063 &  0.178 & 2.129 \\
    & MVSNet~\cite{yao2018mvsnet}  & $0.109^*$  & $1.62^*$  & $8.21^*$  & - & - & - \\
    % & IterMVS~\cite{wang2022itermvs}  & 0.134 & 2.36 & 10.78 & 0.061 & 0.198 & 2.381 & 0.34\\
    & IterMVS~\cite{wang2022itermvs}  & $0.104^*$  & $1.59^*$  & $7.95^*$  & $0.057^*$  & $0.178^*$  & $2.234^*$  \\
    & MVS2D~\cite{yang2022mvs2d}  & $0.132^*$  & $2.05^*$  & $9.82^*$  & $0.058^*$  & $0.176^*$  & $2.277^*$  \\
    & SC-GAN~\cite{wu2019spatial} & - & - & - & 0.063 &  0.178 & 2.129 \\
    & MaGNet~\cite{bae2022multi}  & $0.112^*$  & $1.74^*$  & $9.23^*$  & 0.054 & 0.162 & 2.158 \\
    % & \baseline{Mvs2d-AF}  & \baseline{0.098} & \baseline{1.55} & \baseline{7.86} & \baseline{0.043}  & \baseline{0.135} & \baseline{1.861} & \baseline{72.5}\\
    & \baseline{AFNet} & \baseline{\textbf{0.088}} & \baseline{\textbf{1.41}} & \baseline{\textbf{7.23}} & \baseline{\textbf{0.039}} & \baseline{\textbf{0.121}} & \baseline{\textbf{1.743}} \\
    \bottomrule
  \end{tabular}
  % \vspace{-5pt}
  \caption{Quantitative evaluation on DDAD~\cite{godard2019digging} and KITTI~\cite{geiger2013vision}. Note that the $*$ marks the result reproduced by us using their open-source code, other reported numbers are from the corresponding original papers.}
  \label{tab:DDAD}
  \vspace{-15pt}
\end{table*}

\subsection{Main Results}
\label{sec:4.3}
To demonstrate the outstanding performance of our method, we evaluate AFNet on DDAD~\cite{godard2019digging} and KITTI~\cite{eigen2014depth}. 

\textbf{DDAD~\cite{godard2019digging}.} Since most of classical multi-view and single-view methods are not trained and tested on DDAD, We %apply 
employ the same training scheme to all methods (see %following 
Section~\ref{sec:Implementation details} for details).  Note that all methods have converged well. In the testing, %In particular, 
we evaluate all ring cameras instead of the front-view camera. %, i.e., the final result is the average of six camera views, rather than only testing on the %simplest 
%forward-facing 
%front-view camera. 
Quantitative comparisons are %As 
%shown 
reported in Table \ref{tab:DDAD}, our AFNet %has reached SOTA
achieves the state-of-the-art (SOTA) performance on DDAD.
Compared with %the SOTA
current SOTA methods~\cite{wang2022itermvs} and~\cite{bae2022multi}, our AFNet %has a 15.4\% and 21.4\% improvement on AbsRel error respectively.
can achieve over 15\% improvement on AbsRel error. Qualitative comparisons are shown in Figure~\ref{fig:ddadopt}. The proposed AFNet achieves better results both on dynamic objects and static objects.

\textbf{KITTI~\cite{geiger2013vision}.} KITTI Eigen split~\cite{eigen2014depth} is an %classical 
important benchmark for single-view and multi-view depth estimation. We compare with the state-of-the-art methods on it and show results in Table \ref{tab:DDAD}. Our AFNet achieves 0.039 AbsRel, outperforming recent methods by a large margin. It is worth mentioning that the AbsRel error of AFNet is reduced by 27.8\% compared with the SOTA method~\cite{bae2022multi}.

\subsection{Ablation Study\label{sec:4.5}}
To demonstrate the effectiveness of each %strategy 
component of AFNet,  we ablation on DDAD. Following \cite{yang2022mvs2d},
%we utilize MVS2D~\cite{yang2022mvs2d} and replace its encoder 
we design a two-branch network and employ the ConvNeXt-T~\cite{liu2022convnet} as %the baseline model (denote as Base) in this study on the DDAD~\cite{godard2019digging} test set. 
the backbone. In this section, the following variants are discussed to verify the effectiveness of the proposed AFNet:

\begin{itemize}
\item \textbf{Base}: Baseline is a multi-view depth estimation model~\cite{yang2022mvs2d}, which has the same backbone as ours.
    \item \textbf{Results Fusion}: We add a decoder to the baseline model (Base) for the single-view branch to get single-view prediction and use Ground Truth for supervision, and $\boldsymbol{d}_{fuse}$  is obtained by directly fusing the results of the two branches through two convolutional layers, denoted as Base-RF.
    \item \textbf{Feature Fusion}: We integrate the depth features of the single-view branch into the multi-view branch for complementing semantic and depth cues, denoted as Base-FF.
    \item \textbf{Adaptive Fusion}: Our proposed AF module adaptively selects the most accurate depth between the two branches, demoted as Base-AF.
\end{itemize}

\textbf{Ablation on results fusion.}
The comparison results are shown in Table~\ref{tab:ablation}. Compared with `Base', adding a decoder to the single-view branch and supervising the output can extract more robust features for the epipolar attention, leading to the SqRel error of the multi-view prediction (`Base (Multi)' v.s. `Base-RF (Multi)') reduced by 6.3\%. However, a naive fusion of the single-view and the multi-view branch will drop the final performance, i.e. the accuracy of `Base-RF (Fuse)' is lower than `Base-RF (Multi)'.

\textbf{Effectiveness of the feature fusion.} According to the previous analysis, single-view and multi-view extract feature through different modes, which can be complementary.  Therefore, by introducing the single view depth feature into the multi-view branch, `Base-FF (Multi)' reduces the SqRel error of multi-view branch prediction by 10.7\% compared with `Base-RF (Multi)'. 

\textbf{Effectiveness of the Adaptive Fusion.} As shown in Table~\ref{tab:ablation}, the accuracy of the fusion result `Base-FF (Fuse)' is lower than `Base-FF (Multi)' since the previous method of direct convolution fusion was too crude.
Thus, we propose an adaptive fusion (AF) module to replace this naive fusion way. Comparing `Base-AF (Fuse)' with `Base-FF (Fuse)', using the AF module for results fusion has a 13.4\% performance improvement. In addition, the fusion result `Base-AF (Fuse)' reduces the SqRel error by 2.5\% compared with multi-view branch prediction `Base-AF (Multi)'.
% \begin{table}[t]
%   \centering
%   \tablestyle{4pt}{1.2}
%   \begin{tabular}{c|cccccc}
%     \multirow{2}{*}{Model} & \multicolumn{2}{c}{Single} & \multicolumn{2}{c}{Multi} & \multicolumn{2}{c}{fuse} \\
%      & abs-rel & sq-rel & abs-rel & sq-rel & abs-rel & sq-rel \\ 
%     \shline
%     Base & 0.168 & 2.68 &  0.115 & 1.78 & 0.128 & 1.99 \\
%     Base-s & 0.168 & 2.66 &  0.099 & 1.59 & 0.116 & 1.79 \\
%     Base-2 & 0.101 & 1.64 & 0.099 & 1.58 & 0.103 & 1.66\\
%   \end{tabular}
%   \caption{Results.   Ours is better.SEE THE COMMENTS.}
%   \label{tab:fuse}
% \end{table}

% \subsubsection{Two branches for prediction\label{sec:4.6}}
% \indent We first explore the impact of the two branches on the overall accuracy.  

\begin{table}[t]
  \centering
  % \tablestyle{1.5pt}{1.1}
  \begin{tabular}{c|ccccc}
    \toprule
    Model &RF&FF& Single & Multi & Fuse\\
    \hline
    Base & &&- & 1.90 & - \\
    Base-RF &Conv&&2.68 & 1.78 & 1.99\\
    Base-FF &Conv&\checkmark& \textbf{2.66} & \textbf{1.59} & 1.79\\
    \baseline{Base-AF} &\baseline{Adaptive}&\baseline{\checkmark}& \baseline{\textbf{2.66}} & \baseline{\textbf{1.59}} & \baseline{\textbf{1.55}} \\
    \bottomrule
  \end{tabular}
  \vspace{-8pt}
  \caption{Results of ablation experiments for each strategy in our method on DDAD~\cite{godard2019digging}. Single represents the result of single-view branch prediction, Multi represents the result of multi-view branch prediction, Fuse represents the fusion result $\boldsymbol{d}_{fuse}$. The reported numbers are SqRel error.}
  \label{tab:ablation}
\end{table}

% \subsubsection{Two branches feature fuse\label{sec:4.6}}
% \indent According to the previous analysis, single-view and multi-view predict depth through different modes, which can be complementary. Therefore, we integrate the depth features of the single-view branch into the multi-view branch for complementing semantic and depth cues, denoted as Base-2b-s. As shown in Table \ref{tab:ablation}, compared with Base-2b, Base-2b-s reduce the SqRel error by 10.7\% for multi-view branch prediction.

% \subsubsection{Effectiveness of adaptive fusion  module\label{sec:4.7}}
% We demonstrate the effectiveness of our AF module from three aspects:

% \textbf{Higher accuracy than naive fusion method.} Although the previous series of strategies improved the accuracy, the fusion result $\boldsymbol{d}_{fuse}$ accuracy is lower than multi-view branch, since the previous method of direct convolution fusion was too crude.
% Thus, we propose adaptive fusion (AF) module to replace this crude fusion way, denoted as Base-2b-s-AF.
% This section explores the influence of AF module, two groups of comparison experiments were set up, and AF module was used to replace the simple convolution for fusion on Base-s and Base-2, respectively denoted as Base-s-AF and Base-2-AF. 

\textbf{Ablation on other strategies.} Compared with Base-AF (denoted as BASE for ease of description in this section), our AFNet mainly consists of the following two adjustments: 1) parameter sharing of feature extraction network of single-view and multi-view branch, denoted as BASE-PS; 2) replace Epipolar Attention Module in~\cite{yang2022mvs2d} with cost volume illustrated in Section~\ref{sec:single and multi} to extract matching information, denoted as BASE-PS-cost. As shown in Table \ref{tab:strategy}, compared with BASE, our BASE-PS-cost has a 9.0\% performance improvement for the final result, which is also our final network model, denoted as AFNet. 

\begin{table}[t]
  \centering
  \begin{tabular}{c|ccc}
    \toprule
    Model & Single & Multi & Fuse\\
    \hline
    BASE &2.66 & 1.59 & 1.55 \\
    BASE-PS& 2.62 & 1.53 & 1.49\\
    \hline
    \baseline{\makecell[c]{AFNet \\(BASE-PS-cost)}}& \baseline{2.62} & \baseline{\textbf{1.44}}& \baseline{\textbf{1.41}}\\
    \bottomrule
  \end{tabular}
  \vspace{-8pt}
  \caption{Ablation results on feature extraction network parameter sharing and methods for extracting matching information. The reported numbers are SqRel error.}
  \label{tab:strategy}
  \vspace{-5pt}
\end{table}

\subsection{Discussions}
In this section, we discuss the robustness of AFNet in dynamic regions, on zero-shot datasets, and under noise poses.

\subsubsection{Performance in Dynamic Object Region}

By adaptively fusing monocular depth, our AF module can alleviate the problem that multi-view methods cannot handle dynamic objects. This is because the dynamic object region does not satisfy the projection relation, warping confidence map $\boldsymbol{M}_{w}$ in AF module has certain ability to recognize the dynamic object region, so it can adaptively fuse the single-view result in this region for better prediction. The method of obtaining the mask of the dynamic object region can be found in the supplementary material. As shown in Table \ref{tab:Robust}, our fusion result `AFNet($\boldsymbol{d}_{fuse}$)' has a 21.0\% improvement on SqRel error compared with multi-view prediction `AFNet($\boldsymbol{d}_{m}$)'. We also compare with the SOTA single-view and multi-view fusion method~\cite{bae2022multi}, ~\cite{bae2022multi} improve depth accuracy by fusing single-view depth probability with multi-view geometry, but this kind of multi-view geometry in dynamic object region is inaccurate which could bring into wrong guidance, and it can be seen that our system performs better in the dynamic object region.

\begin{table}[ht]
  \centering
  \begin{tabular}{@{}c|ccc@{}}
    \toprule
    Model & AbsRel$\downarrow$ & SqRel$\downarrow$ & RMSE$\downarrow$ \\
    \hline
    MVS2D~\cite{yang2022mvs2d} & 0.163 & 2.362 & 7.325\\
    MaGNet~\cite{bae2022multi} & 0.169 & 2.500 & 7.783\\
    AFNet($\boldsymbol{d}_{m}$) & 0.158 & 1.879 & 6.024 \\
    AFNet($\boldsymbol{d}_{fuse}$) & \textbf{0.145} & \textbf{1.484} & \textbf{5.675} \\
    % Baseline & 0.125  & 0.156 & 0.176 & 0.182 & 0.172\\
    % \hline
    % \baseline{Mvs2d-AF} & \baseline{0.098}  & \baseline{0.128} & \baseline{0.161} & \baseline{0.169} & \baseline{0.168}\\
    
    % \baseline{AFNet-pose} & \baseline{0.093}  & \baseline{\textbf{0.123}} & \baseline{\textbf{0.142}} & \baseline{\textbf{0.154}} & \baseline{\textbf{0.164}}\\ 
    \bottomrule
  \end{tabular}
  \vspace{-5pt}
  \caption{Performance comparison in dynamic object region on DDAD. AFNet($\boldsymbol{d}_{m}$) denotes the results of multi-view branch.  AFNet($\boldsymbol{d}_{fuse}$) denotes the fusion result.}
  % , which are improved by AF module

  \label{tab:Robust}
\vspace{-15pt}
\end{table}

\begin{table}[h]
  \centering
  \begin{tabular}{@{}c|ccc@{}}
    \toprule
    Model & AbsRel$\downarrow$ & SqRel$\downarrow$ & RMSE$\downarrow$ \\
    \hline
    IterMVS\cite{wang2022itermvs} & 0.123 & 0.056 & 0.332\\
    MVS2D~\cite{yang2022mvs2d} & 0.098 & 0.044 & 0.276\\
    MaGNet~\cite{bae2022multi} & 0.112 & 0.051 & 0.314\\
    AFNet & \textbf{0.091} & \textbf{0.039} & \textbf{0.253} \\
    % Baseline & 0.125  & 0.156 & 0.176 & 0.182 & 0.172\\
    % \hline
    % \baseline{Mvs2d-AF} & \baseline{0.098}  & \baseline{0.128} & \baseline{0.161} & \baseline{0.169} & \baseline{0.168}\\
    
    % \baseline{AFNet-pose} & \baseline{0.093}  & \baseline{\textbf{0.123}} & \baseline{\textbf{0.142}} & \baseline{\textbf{0.154}} & \baseline{\textbf{0.164}}\\ 
    \bottomrule
  \end{tabular}
  \vspace{-5pt}
  \caption{Zero-shot performance on ScanNet~\cite{dai2017scannet}. All the models are trained on the DDAD dataset and tested on ScanNet. The proposed adaptive fusion network shows a better cross dataset generalization ability.}
  \label{tab:genel}
\vspace{-15pt}
\end{table}

\subsubsection{Generalization}
% Current single-view depth estimation models mostly learn priors that do not generalize well. However, the multi-view branch whose depth is obtained mainly by multi-view geometry has better generalization. 
The integration of single-view feature into multi-view branch for complementary and the use of adaptive fusion module to select more accurate depth operations are conducive to network robustness and generalization. To observe the generalization of our network, we evaluated AFNet through cross-dataset testing, i.e. testing our model trained on DDAD directly on ScanNet, as shown in Table \ref{tab:genel}, it shows our model has better performance than the current SOTA methods on ScanNet~\cite{dai2017scannet}.
\vspace{-10pt}

\subsubsection{Robustness under Noise Poses}
Noise is inevitable in various SLAM methods to retrieve pose, but whether the pose is accurate or not greatly affects the accuracy of multi-view depth estimation. Therefore, in practical applications, the robustness of the depth estimation network under noisy poses is critical. 
% A key role of AF module is to improve the robustness of multi-view depth estimation methods under noise poses. 
We provide two kinds of noise to test the robustness of the proposed AFNet, including different levels of synthetic noise and real-world noise generated by SLAM systems. To reflect the anti-noise ability more comprehensively, we added noise to the input pose in the training process of all the methods in Table \ref{tab:Robust1} and Table \ref{tab:slam} to enable the networks to adapt to various modes of accurate poses and noise poses simultaneously.

\textbf{Synthetic noise.} We propose a new robustness testing benchmark to explore the effectiveness of the multi-view methods given noisy poses. One model is required to be evaluated under different noise levels of the pose, which is obtained by converting the relative poses into Euler angles and translations and then multiplying the Euler angles and translations by a disturbance coefficient. Different coefficient corresponds to different noise level, respectively: 1) accurate pose, denoted as $\delta = 0$; 2) the disturbance coefficient of 50\% test samples is $(1+0.01)$, and the remaining 50\% samples is $(1-0.01)$, denoted as $\delta = 0.01$; 3) the disturbance coefficient of 50\% samples is $(1+0.025)$, and the remaining 50\% samples is $(1-0.025)$, denoted as $\delta = 0.025$; 4) the disturbance coefficient of 50\% test samples is $(1+0.05)$, and the remaining 50\% samples is $(1-0.05)$, denoted as $\delta = 0.05$; 5) set input relative pose as identity pose, and the input source images are the same as the reference image, denoted as ID. The mean relative error ($\mu$) and the standard deviation ($\sigma$) are calculated based on all noise pose levels. Finally, a robustness-aware relative error (R-Rel) is defined as $\mu+\sigma$. The lower the better.

\begin{table}
  \centering
  \tablestyle{1.5pt}{1.02}
  \begin{tabular}{@{}c|cccccc@{}}
    \toprule
    Method & $\delta=0$ & $\delta=0.01$ & $\delta=0.025$ & $\delta=0.05$  & ID &R-Rel \\
    \hline
    PMNet~\cite{wang2021patchmatchnet} & 0.144  & 0.235 & 0.354 & 0.382 & 0.246&0.359\\ 
    CasMVS~\cite{gu2020cascade} & 0.131  & 0.165 & 0.195 & 0.215 & 0.181&0.206\\
    MVSNet~\cite{yao2018mvsnet} & 0.112  & 0.160 & 0.189 & 0.208 & 0.177&0.202\\
    MVS2D~\cite{yang2022mvs2d} & 0.133  & 0.159 & 0.179 & 0.184 & 0.173&0.184\\
    IterMVS~\cite{wang2022itermvs} & 0.107  & 0.154 & 0.178 & 0.187  & 0.175&0.189\\
    MaGNet~\cite{bae2022multi} & 0.115 & 0.162 & 0.185 & 0.191 & 0.183 &0.195\\
    % Baseline & 0.125  & 0.156 & 0.176 & 0.182 & 0.172\\
    % \hline
    % \baseline{Mvs2d-AF} & \baseline{0.098}  & \baseline{0.128} & \baseline{0.161} & \baseline{0.169} & \baseline{0.168}\\
    AFNet($\boldsymbol{d}_{m}$) & 0.095 & 0.137 & 0.163 & 0.175 & 0.170 & 0.178\\
    \baseline{AFNet} & \baseline{\textbf{0.092}}  & \baseline{\textbf{0.125}} & \baseline{\textbf{0.155}} & \baseline{\textbf{0.164}} & \baseline{\textbf{0.165}}&\baseline{\textbf{0.168}}\\
    
    % \baseline{AFNet-pose} & \baseline{0.093}  & \baseline{\textbf{0.123}} & \baseline{\textbf{0.142}} & \baseline{\textbf{0.154}} & \baseline{\textbf{0.164}}\\ 
    \bottomrule
  \end{tabular}
  \vspace{-3pt}
  \caption{Performance comparison of AFNet and some state-of-the-art networks under noisy poses on DDAD~\cite{godard2019digging}. The reported numbers are AbsRel error at different settings, $\delta$ represents the intensity of the noise which increases from 0 gradually. ``ID'' means that we set the input relative pose to the identity pose, and the input source images are the same as the reference image. R-rel is the proposed robustness-aware relative error. AFNet($\boldsymbol{d}_{m}$) denotes the results of multi-view branch $\boldsymbol{d}_{m}$, i.e, without AF module.}
  \label{tab:Robust1}
  % \vspace{-5pt}
\end{table}
\begin{figure*}[h]
\centering
\includegraphics[width=0.96\textwidth]{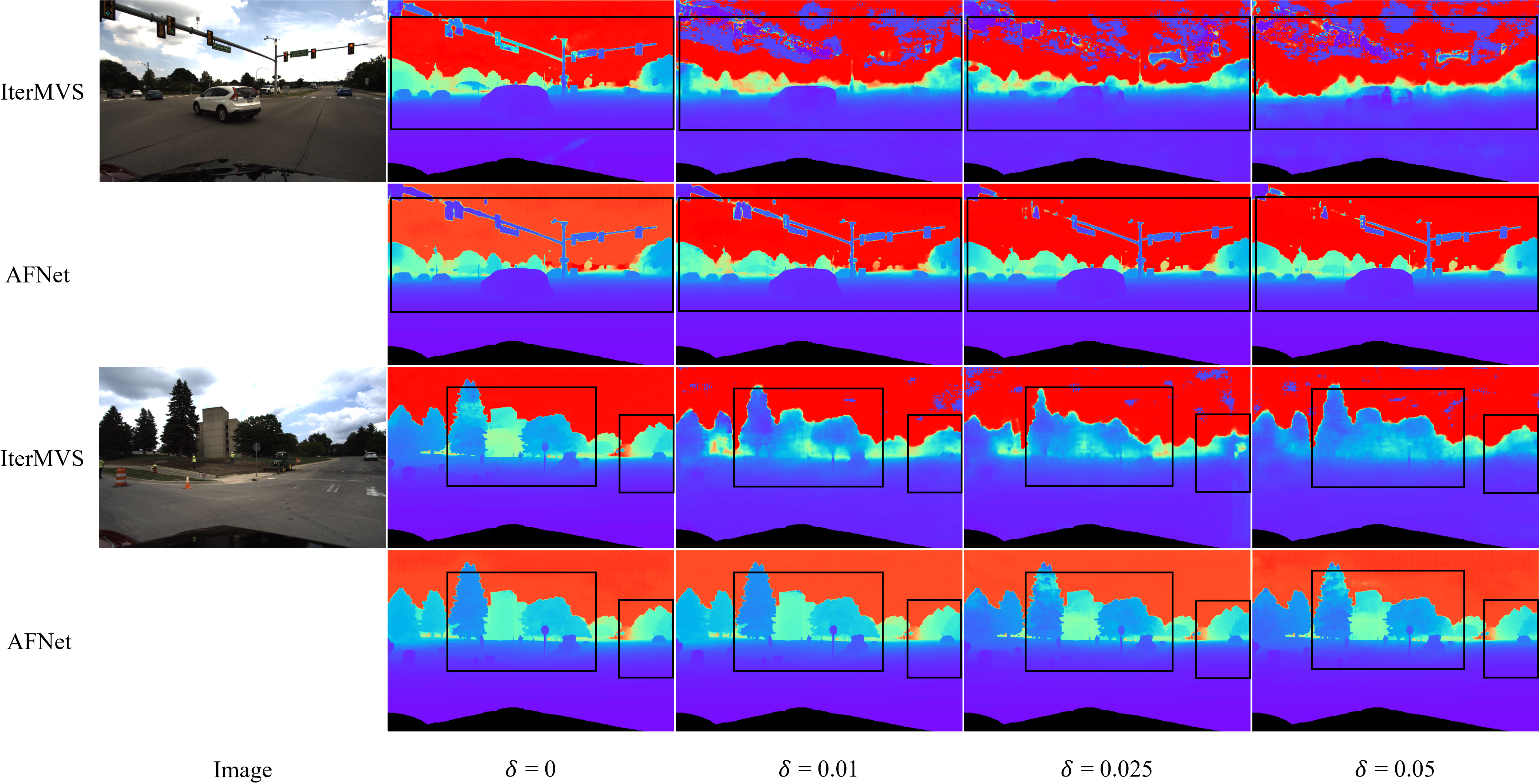} % Reduce the figure size so that it is slightly narrower than the column.
% \vspace{-10pt}
\caption{Visualization comparison results on DDAD~\cite{godard2019digging}. The black boxes show the robustness of our AFNet. With the gradual increases of pose noise, the accuracy of ~\cite{wang2022itermvs} which is mainly based on multi-view matching decreased dramatically, while we remained stable.}
\label{fig:compare}
% \vspace{-15pt}
\end{figure*}

\begin{table}[t]
  \centering
  \tablestyle{1.05pt}{1.01}
  \begin{tabular}{c|ccc|ccc}
    \toprule
    Sequence & \multicolumn{3}{c|}{04} & \multicolumn{3}{c}{05} \\
    \hline
    \multirow{2}{*}{Pose} & \multirow{2}{*}{GT} & ORB2 & ORB1 & \multirow{2}{*}{GT} & ORB2 & ORB1 \\ 
     & & \scriptsize(ATE=0.15)& \scriptsize(ATE=0.98) & & \scriptsize(ATE=0.4)& \scriptsize(ATE=5.6) \\
    \hline
    MoRec\cite{wimbauer2021monorec} & 0.075 & 0.088 & 0.345 & 0.068 & 0.085 & 0.381 \\
    IterMVS\cite{wang2022itermvs} & 0.068 & 0.078 & 0.182 & 0.073 & 0.080 & 0.169 \\
    MaGNet\cite{bae2022multi} & 0.066 & 0.071 &  0.077 & 0.069 & 0.077 & 0.082 \\
    \baseline{AFNet} & \baseline{\textbf{0.059}} & \baseline{\textbf{0.061}} &  \baseline{\textbf{0.064}} & \baseline{\textbf{0.063}} & \baseline{\textbf{0.066}} & \baseline{\textbf{0.070}} \\
    \bottomrule
  \end{tabular}
  % \vspace{-5pt}
  \caption{Performance comparison under Ground Truth poses and SLAM system poses (ORB1 and ORB2 represents the monocular version and stereo version of ORBSLAM2\cite{mur2017orb} respectively) on KITTI~\cite{geiger2013vision} Odometry dataset. ATE represents the absolute trajectory error between the estimated poses and the Ground Truth poses. The reported numbers are AbsRel error.}
  \label{tab:slam}
  % \vspace{-5pt}
\end{table}

 As shown in Table \ref{tab:Robust1}, our AFNet outperforms other classical multi-view methods for the ability to resist input pose noise. When pose has larger noise, the accuracy of classical cost volume-based multi-view methods~\cite{yao2018mvsnet,wang2021patchmatchnet,gu2020cascade} is even much lower than that of single-view methods, such as AdaBins~\cite{bhat2021adabins} and BTS~\cite{lee2019big} whose AbsRel errors are 0.164 and 0.169 as shown in Table \ref{tab:DDAD}. This is because~\cite{yao2018mvsnet,wang2021patchmatchnet,gu2020cascade} obtain depth by regression based on the matching information which is false when poses fail. Some current SOTA multi-view methods also integrate single-view information, such as MVS2d~\cite{yang2022mvs2d} and MaGNet~\cite{bae2022multi}. However, since they do not have the structure to autonomously judge when they can rely on the results of the multi-view methods, thus, when the pose is seriously disturbed, the multi-view branch will introduce wrong guidance.
%  leading to the 
% % they can only forcibly fit this pose degradation situation in the training, 
% accuracy is still lower than that of the single-view methods. 
We overcome this with the proposed adaptive fusion module, AFNet has the highest precision in all noisy pose settings and always remains above the level of the single-view method. There is also an appreciable improvement in robustness-aware relative error (R-Rel) compared with the SOTA fusion system\cite{yang2022mvs2d, bae2022multi}, which is an overall assessment of the robustness of the system. 

% AFNet has a 8.7\% and 13.8\% improvement on R-Rel error respectively compared with the SOTA fusion system \cite{yang2022mvs2d, bae2022multi} respectively.  

Whats more, by comparing the result of the multi-view branch (denoted as AFNet($\boldsymbol{d}_{m}$)) with the result after fusing (denoted as AFNet), it can be seen that the effect of AF module will be more obvious when under noisy poses.
% However, our AFNet overcomes this, and AFnet-Pose can achieve higher accuracy than single-view method even in the case of pose failure through the added pose correction module.

\textbf{Real-world noise poses generated by SLAM systems.}
To more comprehensively evaluate the robustness of our system under noise poses, we tested the depth accuracy under poses obtained by different SLAM systems on KITTI~\cite{geiger2013vision} Odometry dataset. We selected a representative slam system, i.e. ORBSLAM2\cite{mur2017orb}, and used two versions of it to obtain poses respectively, one monocular version (denoted as ORB1) and one stereo version (denoted as ORB2). The monocular version will sometimes crush, and we compare the depth accuracy of the image sequences before crush. The pose noise of the monocular version is larger than that of the stereo version, which can be reflected by the absolute trajectory error. As shown in Table \ref{tab:slam}, our AFNet has the highest accuracy under different poses, and is also the most robust. The results of the remaining sequences in the KITTI Odometry dataset can be found in the supplementary materials.

% \begin{table}[t]
%   \centering
%   \tablestyle{1.5pt}{1.05}
%   \begin{tabular}{@{}c|cccccc@{}}
%     \toprule
%     Method & $\delta=0$ & $\delta=0.01$ & $\delta=0.025$ & $\delta=0.05$  & ID  & R-Rel \\
%     \hline
%     % Baseline & 0.125  & 0.156 & 0.176 & 0.182 & 0.172\\
%     % \hline
%     % \baseline{Mvs2d-AF} & \baseline{0.098}  & \baseline{0.128} & \baseline{0.161} & \baseline{0.169} & \baseline{0.168}\\
%     AFNet & \textbf{0.092}  & 0.125 & 0.155 & 0.164 & 0.165 & 0.168\\ 
%     AFNet-Pose & 0.093  & \textbf{0.123} & \textbf{0.142} & \textbf{0.154} & \textbf{0.164} & \textbf{0.160}\\ 
%     \bottomrule
%   \end{tabular}
%   % \vspace{-8pt}
%   \caption{Ablation results for pose correction module on AbsRel error on DDAD~\cite{godard2019digging}.}
%   \label{tab:pose}
%   \vspace{-10pt}
% \end{table}

% \section{Generalization performance} \label{sec:Generalization}

\section{Conclusion} \label{sec:conclusion}
In this paper, we propose a new multi-view and single-view depth fusion network AFNet for alleviating the defects of the existing multi-view methods, which will fail under noisy poses in real-world autonomous driving scenarios. We propose a new robustness evaluation metric and testing benchmark to explore the effectiveness of the multi-view methods under different noise levels. We fuse the single-view and multi-view depth by the proposed adaptive fusion module, improving the accuracy and robustness of the system. AFNet achieves state-of-the-art performance on both KITTI~\cite{geiger2013vision} and DDAD~\cite{godard2019digging} datasets with accurate pose while also outperforms all other classical multi-view methods on robustness testing benchmark under noisy poses.

\noindent\textbf{Acknowledgement.} This work is supported by National Natural Science Foundation of China (62122029, 62061160490, U20B200007).

{
    \small
    \bibliographystyle{ieeenat_fullname}
    \bibliography{main}
}

\clearpage
\setcounter{page}{1}
\maketitlesupplementary

\begin{table}[t]
  \centering
  \tablestyle{1.5pt}{1.05}
  \begin{tabular}{@{}c|cccccc@{}}
    \toprule
    Method & $\delta=0$ & $\delta=0.01$ & $\delta=0.025$ & $\delta=0.05$  & ID  & R-Rel \\
    \hline
    % Baseline & 0.125  & 0.156 & 0.176 & 0.182 & 0.172\\
    % \hline
    % \baseline{Mvs2d-AF} & \baseline{0.098}  & \baseline{0.128} & \baseline{0.161} & \baseline{0.169} & \baseline{0.168}\\
    AFNet & \textbf{0.092}  & 0.125 & 0.155 & 0.164 & 0.165 & 0.168\\ 
    AFNet-Pose & 0.093  & \textbf{0.123} & \textbf{0.142} & \textbf{0.154} & \textbf{0.164} & \textbf{0.160}\\ 
    \bottomrule
  \end{tabular}
  \caption{Ablation results for pose correction module on AbsRel error on DDAD~\cite{godard2019digging}.}
  \label{tab:pose}
\end{table}

\begin{table*}[t]
  \centering
  \begin{tabular}{c|ccc|ccc|ccc}
    \toprule
    Sequence & \multicolumn{3}{c|}{00} & \multicolumn{3}{c|}{06} & \multicolumn{3}{c}{07} \\
    \hline
    \multirow{2}{*}{Pose} & \multirow{2}{*}{GT} & ORB2 & ORB1 & \multirow{2}{*}{GT} & ORB2 & ORB1 & \multirow{2}{*}{GT} & ORB2 & ORB1\\ 
     & & \scriptsize(ATE=0.92)& \scriptsize(ATE=7.15) & & \scriptsize(ATE=0.69)& \scriptsize(ATE=13.8) & & \scriptsize(ATE=0.48)& \scriptsize(ATE=2.91)\\
    \hline
    MoRec\cite{wimbauer2021monorec} & 0.054 & 0.063 & 0.388 & 0.063 & 0.078 & 0.373 & 0.053 & 0.058 & 0.416 \\
    IterMVS\cite{wang2022itermvs} & 0.058 & 0.067 & 0.114 & 0.043 & 0.052 & 0.093 & 0.064 & 0.075 & 0.128 \\
    MaGNet\cite{bae2022multi} & 0.056 & 0.060 &  0.066 & 0.039 & 0.044 & 0.050 & 0.062 & 0.066 & 0.073  \\
    \baseline{AFNet} & \baseline{\textbf{0.052}} & \baseline{\textbf{0.054}} &  \baseline{\textbf{0.058}} & \baseline{\textbf{0.039}} & \baseline{\textbf{0.041}} & \baseline{\textbf{0.044}} & \baseline{\textbf{0.055}} & \baseline{\textbf{0.058}} & \baseline{\textbf{0.063}} \\
    \bottomrule
  \end{tabular}
  \caption{Performance comparison under Ground Truth poses and SLAM system poses (ORB1 and ORB2 represents the monocular version and stereo version of ORBSLAM2\cite{mur2017orb} respectively) on KITTI~\cite{geiger2013vision} Odometry dataset. ATE represents the absolute trajectory error between the estimated poses and the Ground Truth poses. The reported numbers are AbsRel error.}
  \label{tab:slam}
\end{table*}

\begin{table*}[t]
  \centering
  \begin{tabular}{c|c|ccc|ccc|c}
  \toprule
    \multirow{2}{*}{Type} & \multirow{2}{*}{Model} & \multicolumn{3}{c|}{DDAD~\cite{godard2019digging}}  & \multicolumn{3}{c|}{KITTI~\cite{geiger2013vision}} & \multirow{2}{*}{parm(M) $\downarrow$ }\\
    \cmidrule{3-5} \cmidrule{6-8}
    & & AbsRel$\downarrow$  & SqRel$\downarrow$  & RMSE$\downarrow$  & AbsRel$\downarrow$  & SqRel$\downarrow$  & RMSE$\downarrow$  &  \\
    \hline
    \multirow{2}{*}{\makecell[c]{Single \\ View}} 
    & BTS~\cite{lee2019big} & 0.169 & 2.81 & 11.85 & 0.059 &  0.245 & 2.756 & 112.8 \\ 
    & AdaBins~\cite{bhat2021adabins} & 0.164 & 2.66 & 11.08 & 0.058 & 0.190 & 2.360 & 78.0\\ 
    \hline
    \multirow{4}{*}{\makecell[c]{Fusion \\ Methods}}
    & MVS2D~\cite{yang2022mvs2d}  & 0.132 & 2.05 & 9.82 & 0.058 & 0.176 & 2.277 & 24.4\\
    & MaGNet~\cite{bae2022multi}  & 0.112 & 1.74 & 9.23 & 0.054 & 0.162 & 2.158 & 76.4\\
    & AFNet(MobileNetV2) & 0.091 & 1.45 & 7.28 & 0.040 & 0.124 & 1.751 & \textbf{13.9} \\
    & AFNet(ConvNeXt) & \textbf{0.088} & \textbf{1.41} & \textbf{7.23} & \textbf{0.039} & \textbf{0.121} & \textbf{1.743} & 46.1 \\
    \hline
  \end{tabular}
  \caption{Performance comparison on DDAD~\cite{godard2019digging} and KITTI~\cite{geiger2013vision}, our AFNet has higher accuracy and fewer network parameters.}
  \label{tab:DDAD}
\end{table*}

\section{Robustness under real-world noise poses}
Because the monocular version of ORBSLAM2\cite{mur2017orb} crushes severely on some sequences, we compare it on sequences that perform moderately, and only evaluate the image sequence before it crushes. As shown in Table \ref{tab:slam}, the results of the remaining sequences in the
KITTI Odometry dataset also show the robustness of our AFNet.

\section{Dynamic object region mask}

We claim that our adaptive fusion module can alleviate the problem that multi-view methods cannot handle dynamic objects, so we compared the performance in the region of dynamic objects on the front view camera on DDAD~\cite{godard2019digging}. In order to obtain the mask of the dynamic object region, we first obtain the region  mask1 through instance segmentation that is likely to have dynamic objects, such as cars, pedestrians, bicycles, etc. Then, in the time sequence, the mask1 region of the previous image and the next image are warping to the current image, and the SSIM similarity scores are calculated with the current image. The region with the similarity score less than 0.7 is taken as the final dynamic object region mask.

\section{Pose Correction Module} 
\textbf{Method}. Because the prediction of multi-view branch determines the upper limit of the final accuracy of the system, 
% to further improve the accuracy when the pose is severely interfered, 
we propose the pose correction module to adaptively replace the input noisy pose with the pose predicted by Posenet~\cite{kendall2015posenet} into multi-view branch and AF module for further accuracy improvement. 
Specifically, we input the features $\boldsymbol{{F}}_{i,4}$ extracted from the feature extraction network into the decoder part of Posenet to obtain the predicted Euler angles $\boldsymbol{r}_{i,pred}$ and translation $\boldsymbol{t}_{i,pred}$ between the reference and the $i$-th source cameras. The Euler angles are converted to the rotation matrix $\boldsymbol{R}_{i,pred}$ for warping. Then the source images are warped according to the predicted $\boldsymbol{R, t}$ and the input $\boldsymbol{R, t}$ respectively as in Section 3.3 in paper, denoted as $\left\{\boldsymbol{I}_{i,pred}^{'}\right\}_{i=1}^{n-1}$ and $\left\{\boldsymbol{I}_{i,input}^{'}\right\}_{i=1}^{n-1}$. The difference is that the depth used in this warping is single-view prediction $\boldsymbol{d}_{s}$, since it is not associated with pose. The SSIM similarity scores between reference image $\boldsymbol{I}_{0}$ and warping images $\left\{\boldsymbol{I}_{i,pred}^{'}\right\}_{i=1}^{n-1}$ and $\left\{\boldsymbol{I}_{i,input}^{'}\right\}_{i=1}^{n-1}$ are calculated respectively, and the corresponding $\boldsymbol{R, t}$ with large scores are taken as the input of multi-view branch and the adaptive fusion module.

\textbf{Ablation study}.  To improve the depth accuracy when pose degradation is severe, we propose the pose correction module for AFNet, denoted as AFNet-Pose. As shown in Table \ref{tab:pose}, under different intensities of the pose noise $\delta$, AFNet-Pose has a further improvement when the pose is noisy compared with AFNet, especially $\delta$ = 0.025 has a 8.4\% improvement on AbsRel error. 

\section{Parameter comparison}
In order to prove that the effectiveness of our method is not obtained by parameter stacking, we compare the performance of our method with the current single-view and multi-view fusion methods and classical single-view methods. As shown in Table \ref{tab:DDAD}, our AFNet with ConvNeXt~\cite{liu2022convnet} backbone has the highest accuracy, but the number of parameters is larger than~\cite{yang2022mvs2d}, so we replace the ConvNeXt backbone with a lighter backbone MobileNetV2~\cite{sandler2018mobilenetv2}. It can be seen that our AFNet(MobileNetV2) has the highest accuracy and the lowest number of network parameters compared with other methods.
% WARNING: do not forget to delete the supplementary pages from your submission 
% \input{sec/X_suppl}

\end{document}